\def\year{2021}\relax
\documentclass[letterpaper]{article} % DO NOT CHANGE THIS
\usepackage{aaai21}  % DO NOT CHANGE THIS
\usepackage{times}  % DO NOT CHANGE THIS
\usepackage{helvet} % DO NOT CHANGE THIS
\usepackage{courier}  % DO NOT CHANGE THIS
\usepackage[hyphens]{url}  % DO NOT CHANGE THIS
\usepackage{graphicx} % DO NOT CHANGE THIS
\usepackage{multirow}
\usepackage{subfigure}
\usepackage{amsmath}
\usepackage{amsthm}
\usepackage{amsfonts}
\usepackage{xcolor}
\usepackage{xspace}
\usepackage{graphicx}  % DO NOT CHANGE THIS
\usepackage{natbib}  % DO NOT CHANGE THIS AND DO NOT ADD ANY OPTIONS TO IT
\usepackage{caption} % DO NOT CHANGE THIS AND DO NOT ADD ANY OPTIONS TO IT

\newtheorem{myDef}{Definition}

\newtheorem*{problem}{Problem Statement}

\DeclareMathAlphabet\mathbfcal{OMS}{cmsy}{b}{n}

\newcommand{\eat}[1]{}

\newcommand{\eg}{\emph{e.g.},\xspace}
\newcommand{\ie}{\emph{i.e.},\xspace}

\title{Joint Air Quality and Weather Predictions Based on Multi-Adversarial Spatiotemporal Networks}

\author{Jindong Han\textsuperscript{\rm 1}\thanks{Equal contribution.}, Hao Liu\textsuperscript{\rm 1}\footnotemark[1]\thanks{Corresponding author.}, Hengshu Zhu\textsuperscript{\rm 2}, Hui Xiong\textsuperscript{\rm 3}, Dejing Dou\textsuperscript{\rm 1}\\}

\affiliations{
\textsuperscript{\rm 1}Baidu Research, Beijing, China,  \textsuperscript{\rm 2}Baidu Talent Intelligence Center, Baidu Inc, Beijing, China,  \textsuperscript{\rm 3}Rutgers University, USA \\
\{v\_hanjindong, liuhao30, zhuhengshu, doudejing\}@baidu.com, hxiong@rutgers.edu
}

\begin{document}

\maketitle

\begin{abstract}
Accurate and timely air quality and weather predictions are of great importance to urban governance and human livelihood.
Though many efforts have been made for air quality or weather prediction, most of them simply employ one another as feature input, which ignores the inner-connection between two predictive tasks.
On the one hand, the accurate prediction of one task can help improve another task's performance.
On the other hand, geospatially distributed air quality and weather monitoring stations provide additional hints for city-wide spatiotemporal dependency modeling.
Inspired by the above two insights, in this paper, we propose the \emph{\textbf{M}ulti-\textbf{a}dversarial \textbf{s}patio\textbf{te}mporal \textbf{r}ecurrent \textbf{G}raph \textbf{N}eural \textbf{N}etworks}~(\textbf{MasterGNN}) for joint air quality and weather predictions.   
Specifically, we first propose a heterogeneous recurrent graph neural network to model the spatiotemporal autocorrelation among air quality and weather monitoring stations.
Then, we develop a multi-adversarial graph learning framework to against observation noise propagation introduced by spatiotemporal modeling.
Moreover, we present an adaptive training strategy by formulating multi-adversarial learning as a multi-task learning problem.
Finally, extensive experiments on two real-world datasets show that MasterGNN achieves the best performance compared with seven baselines on both air quality and weather prediction tasks.
\end{abstract}

\section{Introduction}

\begin{figure}
  \centering
  \vspace{-4pt}
  \includegraphics[width=8cm]{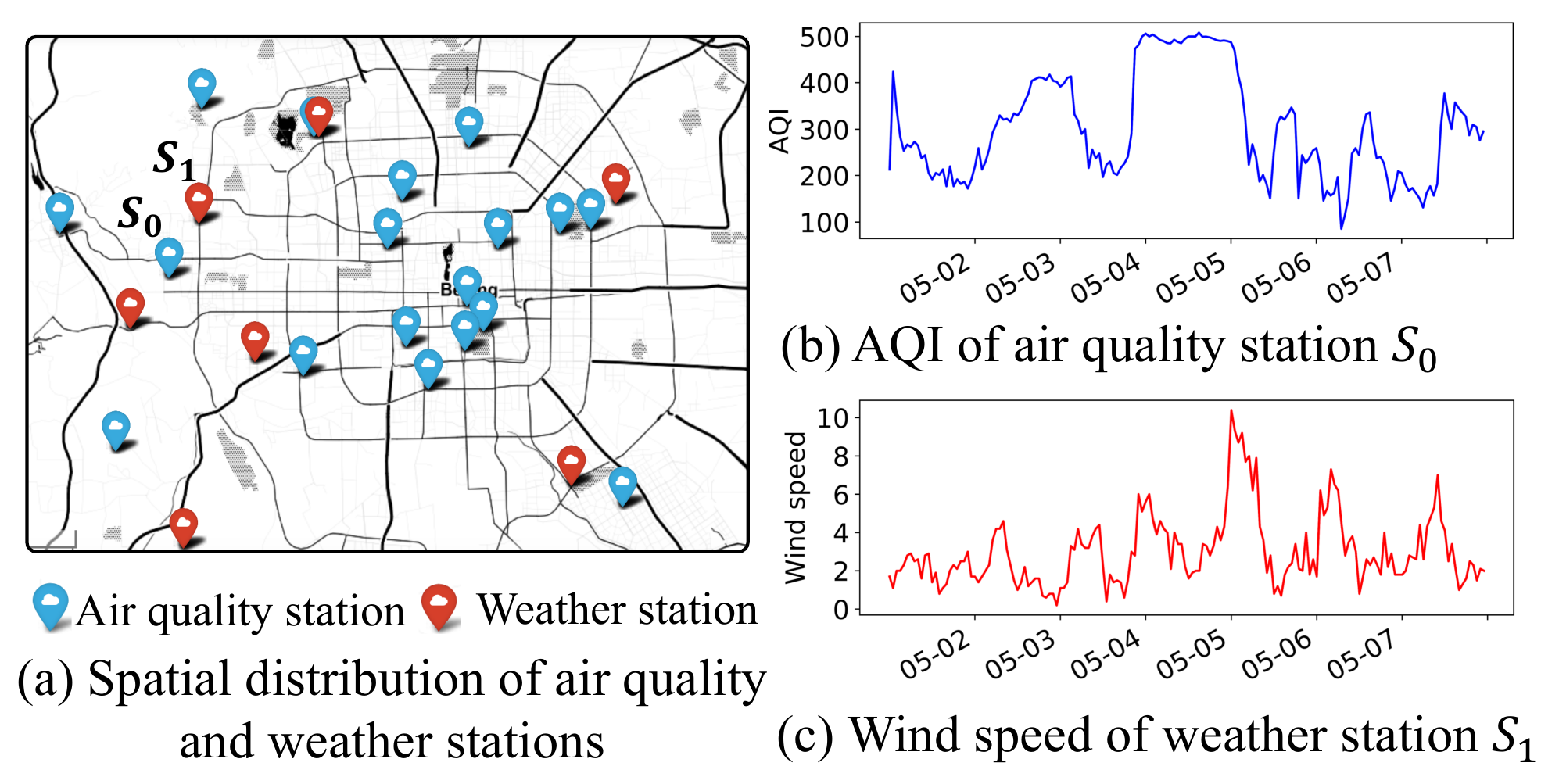}
  \vspace{-5pt}
  \caption{
  Spatial distribution of air quality and weather monitoring stations in Beijing. Two types of stations are monitoring exclusively different but correlated air quality and weather conditions in different city locations.}\label{dist}
  \vspace{-7mm}
\end{figure}

% Two types of stations are monitoring exclusively different but correlated air quality and weather conditions in different city locations.

%Air quality and weather prediction aim at forecasting the future concentration of atmospheric elements~(\eg PM2.5, temperature, and humidity), which are two essential tasks in environmental science. 
%Air quality and weather prediction are two fundamental atmospheric tasks in environmental science, which aim at forecasting the future concentration of atmospheric elements~(\eg PM2.5, temperature, and humidity).

%As two fundamental atmospheric tasks in environmental science, the goal of air quality and weather prediction is to forecast the air quality~(\eg PM2.5, PM10) and weather conditions~(\eg temperature, and humidity) in the next period, respectively.
With the rapid development of the economy and urbanization, people are increasingly concerned about emerging public health risks and environmental sustainability.
%, environmental economics, 
As reported by the World Health Organization~(WHO), air pollution is the world's largest environmental health risk~\cite{campbell2019climate}, and the changing weather profoundly affects the local economic development and people's daily life~\cite{baklanov2018urban}.
Thus, accurate and timely air quality and weather predictions are of great importance to urban governance and human livelihood.
In the past years, massive sensor stations have been deployed for monitoring air quality~(\eg PM2.5 and PM10) or weather conditions~(\eg temperature and humidity) and many efforts have been made for air quality and weather predictions~\cite{liang2018geoman,yi2018deep,wang2019deep}.

%However, existing studies on air quality and weather prediction 
%, by collectively modeling spatiotemporal correlation among historical air quality or weather conditions~\cite{zheng2013u,yi2018deep,wang2019deep}.
%with considering various urban contextual factors~(\eg POI distributions, traffics)~
%Recent studies uncover the air quality and weather conditions are mutually influenced~\cite{ding2017air,hong2020weakening}.
%For example, the stronger wind can promote the dispersion and transport of air pollutants.
%The aerosol pollutants~(\eg PM2.5 and PM10) also can reversely impact the local climate~(\eg temperature and humidity) by scattering or absorbing incoming radiation.
However, existing air quality and weather prediction methods are designated for either air quality or weather prediction, perhaps with one another as a side input~\cite{zheng2015forecasting,yi2018deep}, but overlook the intrinsic connections and interactions between two tasks.
Different from existing studies, this work is motivated by the following two insights. First, air quality and weather predictions are two highly correlated tasks and can be mutually enhanced.
For example, accurately predicting the regional wind condition can help model the future dispersion and transport of air pollutants~\cite{ding2017air}.
As another example, modeling the future concentration of aerosol pollutants~(\eg PM2.5 and PM10) also can help predict the local climate~(\eg temperature and humidity), since aerosol elements can scatter or absorb the incoming radiation~\cite{hong2020weakening}.
Second, as illustrated in Figure~\ref{dist}, the geospatially distributed air quality and weather monitoring stations provide additional hints to improve both predictive tasks.
The air quality and weather condition variations in different city locations reflect the urban dynamics and can be exploited to improve spatiotemporal autocorrelation modeling.
%The observed variation of air quality and weather conditions in different city locations can be exploited for more accurate dynamic spatiotemporal autocorrelation modeling.
%can enhance the modeling of city-wide spatiotemporal dynamics.
%The observations of one kind monitoring stations can be exploited as additional information for predicting the future conditions of the other kind of monitoring stations, by better modeling the city-wide spatiotemporal dynamics.
%exploited explored

%based on historical air quality and weather observations, as well as various urban contextual factors.
%Such mixed distributed stations can be jointly exploited to enhance spatiotemporal urban dynamic sensing and improve air quality and weather prediction.
%geo-distributed air quality and weather stations are monitoring exclusively heterogeneous observations, which provide additional hints of future atmospheric condition variation.
%Besides, the interaction among air quality and weather monitoring stations are conditioned on various urban contextual factors~(\eg tall builidings ).
In this work, we investigate the joint prediction of air quality and weather conditions by explicitly modeling the correlations and interactions between two predictive tasks.
However, two major challenges arise in achieving this goal.
(1) \emph{Observation heterogeneity}.
As illustrated in Figure \ref{dist}, the geo-distributed air quality and weather stations are heterogeneous spatial objects that are monitoring exclusively different atmospheric conditions. 
Existing methods~\cite{zheng2013u,wang2019deep} are initially designed to model homogeneous spatial objects~(\ie either air quality or weather stations), which are not suitable for joint air quality and weather predictions.
Therefore, the first challenge is how to capture spatiotemporal autocorrelations among heterogeneous monitoring stations to mutually benefit air quality and weather prediction.
(2) \emph{Compounding observation error vulnerability}.
In practice, observations reported by monitoring stations are often noisy due to the sensor error and environmental interference~\cite{yi2016st}.
However, most existing prediction models~\cite{zheng2015forecasting,liang2018geoman,yi2018deep} rely on spatiotemporal dependency modeling between stations, which is susceptible to local perturbations and noise propagation~\cite{zugner2018adversarial,bengio2015scheduled}.
More severely, jointly modeling spatiotemporal autocorrelation among air quality and weather monitoring stations will further accumulate errors from both spatial and temporal domains.
As a result, it is challenging to learn robust representations to resist compounding observation error for joint air quality and weather predictions.
%and lead to unsatisfied prediction accuracy.

%Different from existing methods for homogeneous spatial objects
%In particular, multiple microscopic discriminators learn to distinguish adversarial examples from both spatial and temporal domains, and macroscopic discriminators learn to distinguish adversarial examples constructed from a holistic view.
%Moreover, to stabilize the adversarial training, we reformulate the multi-discriminator loss minimization as a multi-objective optimization problem and propose a hierarchical hyper-volume maximization strategy to enforce the optimizer weigh on poorly performed discriminators. Our contributions are summarized as follows.
To tackle the above challenges, we propose the \emph{\textbf{M}ulti-\textbf{a}dversarial \textbf{s}patio\textbf{te}mporal \textbf{r}ecurrent \textbf{G}raph \textbf{N}eural \textbf{N}etworks}~(\textbf{MasterGNN}) for robust air quality and weather joint predictions.
Specifically, we first devise a heterogeneous recurrent graph neural network to simultaneously incorporate spatial and temporal autocorrelations among heterogeneous stations conditioned on dynamic urban contextual factors~(\eg POI distributions, traffics).
%The obtained station representations incorporate both non-Euclidean spatial dependencies and dynamic temporal dependencies among heterogeneous stations under various contextual factors. 
Then, we propose a multi-adversarial learning framework to against noise propagation from both microscopic and macroscopic perspectives.
By proactively simulating perturbations and maximizing the divergence between target and fake observations, multiple discriminators dynamically regularize air quality and weather station representations to resist the propagation of observation noises~\cite{durugkar2016generative}.
Moreover, we introduce a multi-task adaptive training strategy to improve the joint prediction performance by automatically balancing the importance of multiple discriminators. %to stabilize multi-adversarial learning,
Finally, we conduct extensive experiments on two real-world datasets collected from Beijing and Shanghai, and the proposed model consistently outperforms seven baselines on both air quality and weather prediction tasks.
%conduct extensive experiments on two real-world datasets collected from Beijing and Shanghai. The proposed model consistently outperforms seven baselines on both air quality and weather prediction tasks.
\eat{
To sum up, our main contributions are as follows:
\begin{itemize}
	\item We study the novel air quality and weather co-prediction problem by exploiting the inner-connection between air quality and weather conditions.
	\item We propose a heterogeneous recurrent graph neural network to capture the spatiotemporal autocorrelation among heterogeneous monitoring stations.
	\item We introduce a multi-adversarial learning framework coupled with a multi-objective optimization strategy to resist observation noises.
	\item We conduct extensive experiments on two real-world datasets collected from Beijing and Shanghai. The proposed model outperforms seven baselines by at least 10.7\% and 5.5\% on air quality and weather prediction tasks, respectively.
\end{itemize}
}
\section{Preliminaries}

%In this section, we first introduce some important definitions and then formalize the air quality and weather co-prediction problem we aim to investigate.

%\TODO{AQI and Meteorological readings.}

Consider a set of monitoring stations $S=S^a\cup S^w$, where $S^a=\{s^a_i\}^m_{i=1}$ and $S^w=\{s^w_j\}^n_{j=1}$ are respectively air quality and weather station sets.
Each station $s_i \in S$ is associated with a geographical location $l_i$~(\ie latitude and longitude) and a set of time-invariant contextual features $\mathbf{c}_i \in \mathbf{C}$.
%~(\eg POI distribution~\cite{liu2019hydra} and nearby road network density~\cite{zheng2013u}). 
%We first define the time varying observations.

\begin{myDef}
\textbf{Observations}. Given a monitoring station $s_i \in S$, the observations of $s_i$ at time step $t$ are defined as $\mathbf{x}^{t}_i$, which is a vector of air-quality or weather conditions, depending on the station type. 
\end{myDef}

Note that the observations of two types of monitoring stations are different~(\eg PM2.5 and CO in air quality stations, while temperature and humidity in weather stations) and the observation dimensionality of the air quality station and the weather station are also different. 
We use $\mathbfcal{X}=\{\mathbf{X}^{a,t}\}^T_{t=1} \cup \{\mathbf{X}^{w,t}\}^T_{t=1}$ to denote time-dependent observations of all stations in a time period $T$, use $\mathbf{X}^{a,t}=\{\mathbf{x}^{a,t}_1, \mathbf{x}^{a,t}_2, \dots, \mathbf{x}^{a,t}_{m}\}$ and $\mathbf{X}^{w,t}=\{\mathbf{x}^{w,t}_1, \mathbf{x}^{w,t}_2, \dots, \mathbf{x}^{w,t}_{n}\}$ to respectively denote observations of air quality and weather stations at time step $t$. We use $\mathbf{X}_i=\{\mathbf{x}^{1}_i, \mathbf{x}^{2}_i, \dots, \mathbf{x}^{T}_i\}$ to denote all observations of station $s_i$.
% which is a special case of multivariate time series.
In the following, without ambiguity, we will omit the superscript and subscript.

\eat{
\begin{myDef}
\textbf{Heterogeneous Station Graph}. A heterogeneous station graph (HSG) is defined as $\mathcal{G}=\{\mathcal{V},\mathcal{E}, \phi \}$, where $\mathcal{V}=S$ is the set of monitoring stations, $\phi$ is a mapping set indicates the type of each station, and $\mathcal{E}$ is a set of edges indicating the connectivity among monitoring stations, defined as
\begin{equation}
e_{ij}=\left\{
\begin{array}{lr}
1,\quad dist(s_i,s_j)<\epsilon\\
0,\quad otherwise
\end{array}
\right.,
\end{equation}
where $dist(\cdot,\cdot)$ is the spherical distance between station $s_i$ and station $s_j$, $\epsilon$ is a distance threshold.
\end{myDef}
}

\begin{problem}
\textbf{Joint air quality and weather predictions}. 
Given a set of monitoring stations $S$, contextual features $\mathbf{C}$, and historical observations $\mathbfcal{X}$, our goal at a time step $t$ is to simultaneously predict air quality and weather conditions for all $s_i \in S$ over the next $\tau$ time steps,
\begin{equation}
(\hat{\mathbf{Y}}^{t+1},\hat{\mathbf{Y}}^{t+2},...,\hat{\mathbf{Y}}^{t+\tau}) \leftarrow \mathcal{F}(\mathbfcal{X}, \mathbf{C}),
\end{equation}
where $\hat{\mathbf{Y}}^{t+1}=\hat{\mathbf{Y}}^{a,t+1} \cup \hat{\mathbf{Y}}^{w,t+1}$ is the estimated target observations of all stations at time step $t+1$, and $\mathcal{F}(\cdot)$ is the mapping function we aim to learn. Note that $\hat{\mathbf{y}}^{t+1}_i \in \hat{\mathbf{Y}}^{t+1}$ is also station type dependent, \ie air quality or weather observations for corresponding stations.
\end{problem}

\section{Methodology}
%\subsection{Model Intuitions}
MasterGNN simultaneously makes air quality and weather predictions based on the following intuitions.

\textbf{Intuition 1: Heterogeneous spatiotemporal autocorrelation modeling}. The geo-distributed air quality and weather monitoring stations provide additional spatiotemporal hints for air quality and weather prediction. The model should be able to simultaneously incorporate spatial and temporal autocorrelations of such heterogeneous monitoring stations for joint air quality and weather prediction.

\textbf{Intuition 2: Robust representation learning}. 
The spatiotemporal autocorrelation based joint prediction model is vulnerable to observation noises and suffers from compounding propagation errors. The model should be robust to error propagation from both spatial and temporal domains.
%The observation noise may propagate and accumulate during spatiotemporal autocorrelation modeling, leading to unsatisfied prediction accuracy. 

\textbf{Intuition 3: Adaptive model training}. Learning robust representations introduces extra optimization objectives, which guiding the predictive model to approximate the underlying data distribution from different semantic aspects. The model should be able to balance the optimization of multiple objectives for optimal prediction adaptively. 

\subsection{Framework Overview}

\begin{figure}[t]
  \centering
  %\vspace{-0.1cm}
  \includegraphics[width=7.5cm]{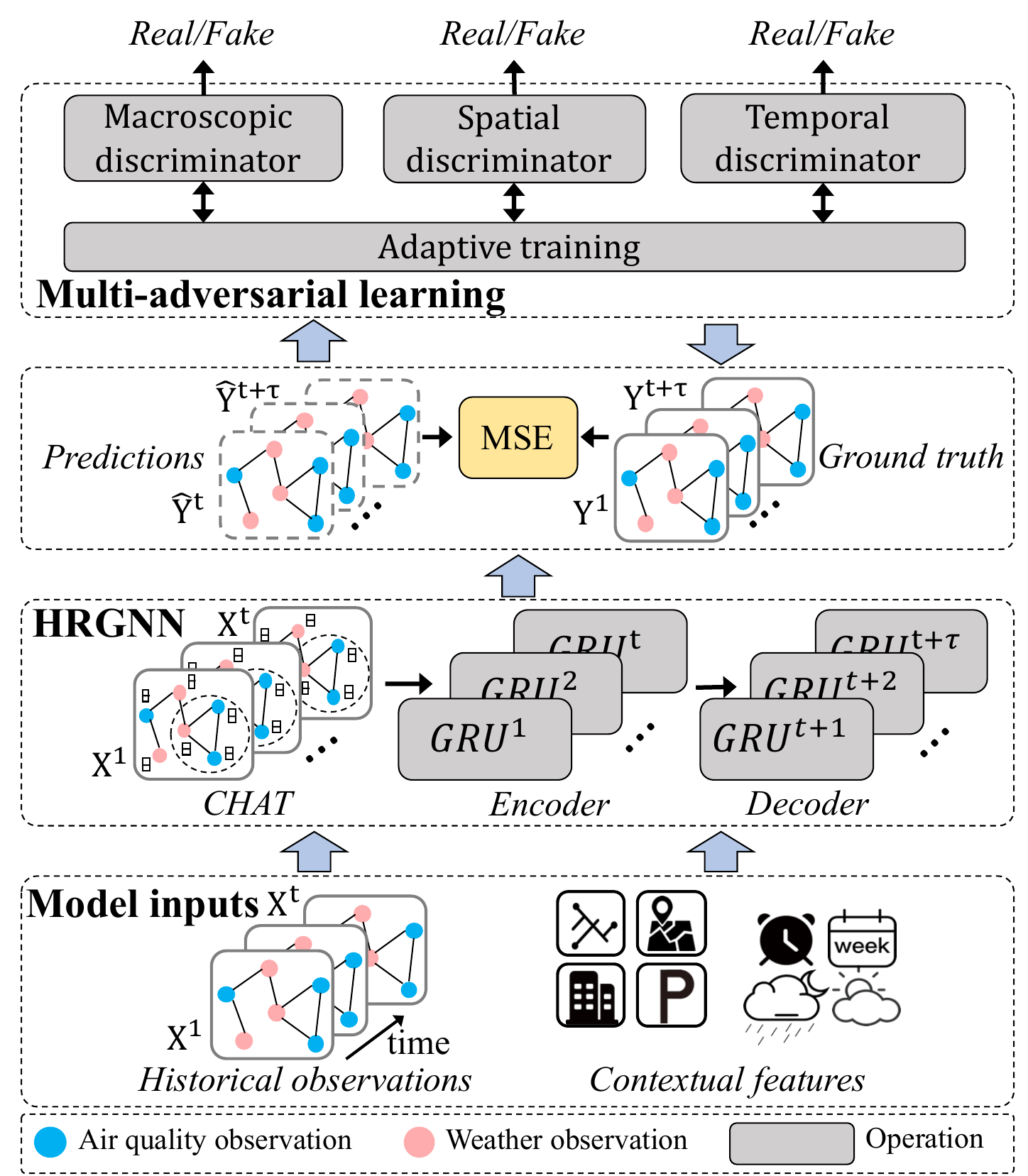}
  \vspace{-0.1cm}
  \caption{An overview of MasterGNN.}\label{framework}
  \vspace{-0.6cm}
\end{figure}

%By taking historical observations and contextual features of all stations as input, our model predicts air quality observations and weather observations for each station in the next $ \tau $ time step.
Figure \ref{framework} shows the architecture of our approach, including three major tasks: 
(i) modeling the spatiotemporal autocorrelation among air quality and weather stations for joint prediction; (ii) learning robust station representations via multi-adversarial learning; (iii) exploiting adaptive training strategy to ease the model learning.
Specifically, in the first task, we propose a \emph{Heterogeneous Recurrent Graph Neural Network}~(HRGNN) to jointly incorporate spatial autocorrelation between heterogeneous monitoring stations and past-current temporal autocorrelation of each monitoring station.
In the second task, we develop a \emph{Multi-Adversarial Learning} framework to resist the propagation of observation noises via (i) microscopic discriminators against adversarial attacks respectively from the spatial and temporal domain, and (ii) macroscopic discriminator against adversarial attacks from a global view of the city.
In the third task, we introduce a \emph{Multi-Task Adaptive Training} strategy to automatically balance the optimization of multiple discriminative losses in multi-adversarial learning.
%we introduce the \emph{Hierarchical Hyper-Volume Maximization} strategy by
%formulating the multi-adversarial training as a multi-objective optimization problem.
%and devise a \emph{Hierarchical Hyper-Volume Maximization} strategy to stabilize the multi-adversarial training.
%formulate the multi-adversarial graph learning as a Multi-objective Optimization problem, where the loss signal of each discriminator is regarded as an independent objective. We devise a \emph{hierarchical hyper-volume maximization} strategy to adaptively balance the convergence of multiple discriminators and further stabilize the adversarial training.
%simultaneously minimize multiple losses to facilitate model convergence.

\subsection{Heterogeneous Recurrent Graph Neural Network}

%In this section, we present the heterogeneous recurrent graph neural network for joint air quality and weather prediction.
%We first present HRGNN in detail.

\subsubsection{The base model.}
We adopt the encoder-decoder architecture~\cite{zhang2020tkde,zhang2020semi} as the base model.
Specifically, the encoder step projects the historical observation sequence of all stations $\mathbfcal{X}$ to a hidden state $\mathbf{H}^t=f_{encoder}(W\mathbfcal{X}+b)$, where $f_{encoder}(\cdot)$ is parameterized by a neural network.
%a recurrent neural network~(RNN) such as Gated Recurrent Units (GRU) and Long short-term memory~(LSTM). 
In the decoder step, we employ another neural network $f_{decoder}(\cdot)$ to generate air quality and weather predictions $(\hat{\mathbf{Y}}^{t+1},\hat{\mathbf{Y}}^{t+2},\dots,\hat{\mathbf{Y}}^{t+\tau})=f_{decoder}(\mathbf{H}^t, \mathbf{C})$, where $\tau$ is the future time steps we aim to predict, and $\mathbf{C}$ are contextual features~\cite{liu2020incorporating}.
Similar with conventional air quality or weather prediction models~\cite{liang2018geoman,li2020weather}, the heterogeneous recurrent graph neural network aims to minimize the \emph{Mean Square Error}~(MSE) between the target observations and predictions,
\begin{equation}\label{equ:mse}
\mathcal{L}_g=\frac{1}{\tau}\sum^{\tau}_{i=1}||\hat{\mathbf{Y}}^{t+i}-\mathbf{Y}^{t+i}||^2_2.
\end{equation}
%$\mathbf{Y}^{t+i}$ and $\hat{\mathbf{Y}}^{t+i}$ are ground truth target observations and predicted target observations of all stations $S$ at time step $t+i$. 

\subsubsection{Incorporating spatial autocorrelation.}
In the spatial domain, the regional concentration of air pollutants and weather conditions are highly correlated and mutually influenced.
Inspired by the recent variants~\cite{wang2019heterogeneous,zhang2019heterogeneous,liu2021vldb} of graph neural network on handling non-Euclidean semantic dependencies on heterogeneous graphs, we devise a context-aware heterogeneous graph attention block (CHAT) to model spatial interactions between heterogeneous stations.
We first construct the heterogeneous station graph to describe the spatial adjacency of each station.

\begin{myDef}
\textbf{Heterogeneous station graph}. A heterogeneous station graph (HSG) is defined as $\mathcal{G}=\{\mathcal{V},\mathcal{E}, \psi \}$, where $\mathcal{V}=S$ is the set of monitoring stations, $\psi$ is a mapping function indicates the type of each station, and $\mathcal{E}$ is a set of directed edges indicating the connectivity among monitoring stations, defined as
\begin{equation}
e_{ij}=\left\{
\begin{array}{lr}
1,\quad d_{ij}<\epsilon\\
0,\quad otherwise
\end{array}
\right.,
\end{equation}
where $d_{ij}$ is the spherical distance between station $s_i$ and station $s_j$, $\epsilon$ is a distance threshold.
\end{myDef}

Due to the heterogeneity of monitoring stations, there are two types of vertices~(air quality station $s^a$, weather station $s^w$) and four types of edges, \ie $\Psi=\{s^a$-$s^a$, $s^w$-$s^w$, $s^a$-$s^w$, $s^w$-$s^a\}$. 
%We use $r \in \Psi$ to denote a particular edge type.
We use $\psi(i)$ to denote the station type of $s_i$, and $r \in \Psi$ to denote the semantic type of each edge.

Formally, given an observation $\mathbf{x}_i$ at a particular time step, we first devise a type-specific transformation layer to project heterogeneous observations into unified feature space,
%\begin{equation}
$\widetilde{\mathbf{x}}_i = \mathbf{W}^{\psi(i)} \mathbf{x}_i$,
%\end{equation}
where $\widetilde{\mathbf{x}}_i\in\mathcal{R}^d$ is a low-dimensional embedding vector, $\mathbf{W}^{\psi(i)}\in\mathcal{R}^{|\mathbf{x}_i|\times d}$ is a learnable weighted matrix shared by all monitoring stations of type $\psi(i)$.

%Consider an observation $\mathbf{x}_i$ and a specific edge type $r$,
Then, we introduce a type-dependent attention mechanism to quantify the non-linear correlation between homogeneous and heterogeneous stations under different contexts. Given a station pair $(s_i,s_j)$ which are connected by an edge of type $r$, the attention score is defined as
\begin{equation}
a^r_{ij} =\frac{Attn(\widetilde{\mathbf{x}}_i,\widetilde{\mathbf{x}}_j,\mathbf{c}_i,\mathbf{c}_j,d_{ij})}{\sum_{k\in\mathcal{N}^r_{i}}Attn(\widetilde{\mathbf{x}}_i,\widetilde{\mathbf{x}}_k,\mathbf{c}_i,\mathbf{c}_k,d_{ik})},
\end{equation}
where $Attn(\cdot)$ is a concatenation based attention function, $\mathbf{c}_i,\mathbf{c}_j\in\mathbf{C}$ are contextual features of station $s_i$ and $s_j$, $d_{ij}$ is the spherical distance between $s_i$ and $s_j$, and $\mathcal{N}^r_i$ is the set of type-specific neighbor stations of $s_i$ in $\mathcal{G}$.
Based on $a^r_{ij}$, we define the context-aware heterogeneous graph convolution operation to update the type-wise station representation,
\begin{equation}
\widetilde{\mathbf{x}}_i^{r\prime} = GConv(\widetilde{\mathbf{x}}_{i}, r) = \sigma(\sum_{j\in \mathcal{N}^r_i}\alpha^r_{ij}\mathbf{W}^r\widetilde{\mathbf{x}}_j),
\end{equation}
where $\widetilde{\mathbf{x}}^{r\prime}_{i}$ is the aggregated node representation based on edge type $r$, $\sigma$ is a non-linear activation function, $\mathbf{W}^r\in\mathcal{R}^{d \times d}$ is a learnable weighted matrix shared over all edges of type $r$.
%For $v_i$, there are two types of neighbors, homogeneous stations and heterogeneous stations, 
Finally, we obtain the updated station representation of $s_i$ by concatenating type-specific representations,
\begin{equation}
\widetilde{\mathbf{x}}^{\prime}_i = ||_{r\in\Psi}GConv(\widetilde{\mathbf{x}}_{i}, r),
\end{equation}
where $||$ is the concatenation operation.
Note that we can stack $l$ graph convolution layers to capture the spatial autocorrelation between $l$-hop heterogeneous stations.

\eat{
Consider an air quality observation $\mathbf{x}^a_i$ and a weather observation $\mathbf{x}^w_j$ at a particular time step, we first employ a type-specific linear transformation layer to project heterogeneous observations into a unified feature space,
\begin{equation}
\mathbf{h}^a_i = \mathbf{W}^a \mathbf{x}^a_i,~~\mathbf{h}^w_j = \mathbf{W}^w \mathbf{x}^w_j,
\end{equation}
where $\mathbf{h}^a_i\in\mathcal{R}^d$ and $\mathbf{h}^w_j\in\mathcal{R}^d$ are low-dimensional embedding vectors, $\mathbf{W}^a\in\mathcal{R}^{|\mathbf{x}^a_i|\times d}$ and $\mathbf{W}^w\in\mathcal{R}^{|\mathbf{x}^w_j|\times d}$ are learnable weighted matrices shared by air quality and weather monitoring stations, respectively.

We use $\psi_i$ to denote the node type of $v_i$, and $\psi_{ij}$ to denote the edge type from $v_i$ to $v_j$.

Consider an atmospheric observation $\mathbf{x}_i$ at a particular time step, we first employ a type-specific linear transformation layer to project heterogeneous observations into a unified feature space,
\begin{equation}
\mathbf{h}_i = \mathbf{W}_{\psi_i} \mathbf{x}_i,
\end{equation}
where $\mathbf{h}_i\in\mathcal{R}^d$ is a low-dimensional embedding vector, $\mathbf{W}_{\psi_i}\in\mathcal{R}^{|\mathbf{X}_i|\times d}$ is a learnable weighted matrix shared by all same-typed monitoring stations.
}

\subsubsection{Incorporating temporal autocorrelation.}
%The sequence-to-sequence model is designated for homogeneous sequence data.
In the temporal domain, the air quality and weather conditions also depend on previous observations.
We further extend the Gated Recurrent Units (GRU), a simple variant of recurrent neural network~(RNN), to integrate the temporal autocorrelation among heterogeneous observation sequences.
Consider a station $s_i$, given the learned spatial representation $\widetilde{\mathbf{x}}^t_i$ at time step $t$, we denote the hidden state of $s_i$ at $t-1$ and $t$ as $\mathbf{h}^{t-1}_i$ and $\mathbf{h}^t_i$, respectively. The temporal autocorrelation between  $\mathbf{h}^{t-1}_i$ and $\mathbf{h}^t_i$ is modeled by
\begin{equation}
\left\{
\begin{aligned}
    &\mathbf{h}^t_i=(1-\mathbf{z}^t_i)\odot \mathbf{h}^{t-1}_i+\mathbf{z}^t_i\odot \widetilde{\mathbf{h}}^t_i\\
    &\mathbf{r}^{t}_i = \sigma{(\mathbf{W}^{\psi(i)}_r[\mathbf{h}^{t-1}_i\parallel\widetilde{\mathbf{x}}^{t}_i]+\mathbf{b}^{\psi(i)}_r)}\\
    &\mathbf{z}^{t}_i = \sigma(\mathbf{W}^{\psi(i)}_z[\mathbf{h}^{t-1}_i\parallel\widetilde{\mathbf{x}}^{t}_i]+\mathbf{b}^{\psi(i)}_z)\\
    &\widetilde{\mathbf{h}}^{t}_i = \tanh(\mathbf{W}^{\psi(i)}_{\widetilde{h}}[\mathbf{r}^{t}_i\odot\mathbf{h}^{t-1}_i\parallel\widetilde{\mathbf{x}}^t_i]+\mathbf{b}^{\psi(i)}_{\widetilde{h}})
\end{aligned},
\right.
\end{equation}
where $\mathbf{o}^t_i$, $\mathbf{z}^t_i$ denote reset gate and update gate at time stamp t, $W_{o}^{\psi(i)}$, $W_{z}^{\psi(i)}$, $W_{\widetilde{h}}^{\psi(i)}$, $b_{o}^{\psi(i)}$, $b_{z}^{\psi(i)}$, $b_{\widetilde{h}}^{\psi(i)}$ are trainable parameters shared by all same-typed monitoring stations, and $\odot$ represents the Hadardmard product. 

\subsection{Multi-Adversarial Learning}

\begin{figure}[t]
  \centering
  %\vspace{-0.1cm}
  \includegraphics[width=8.5cm]{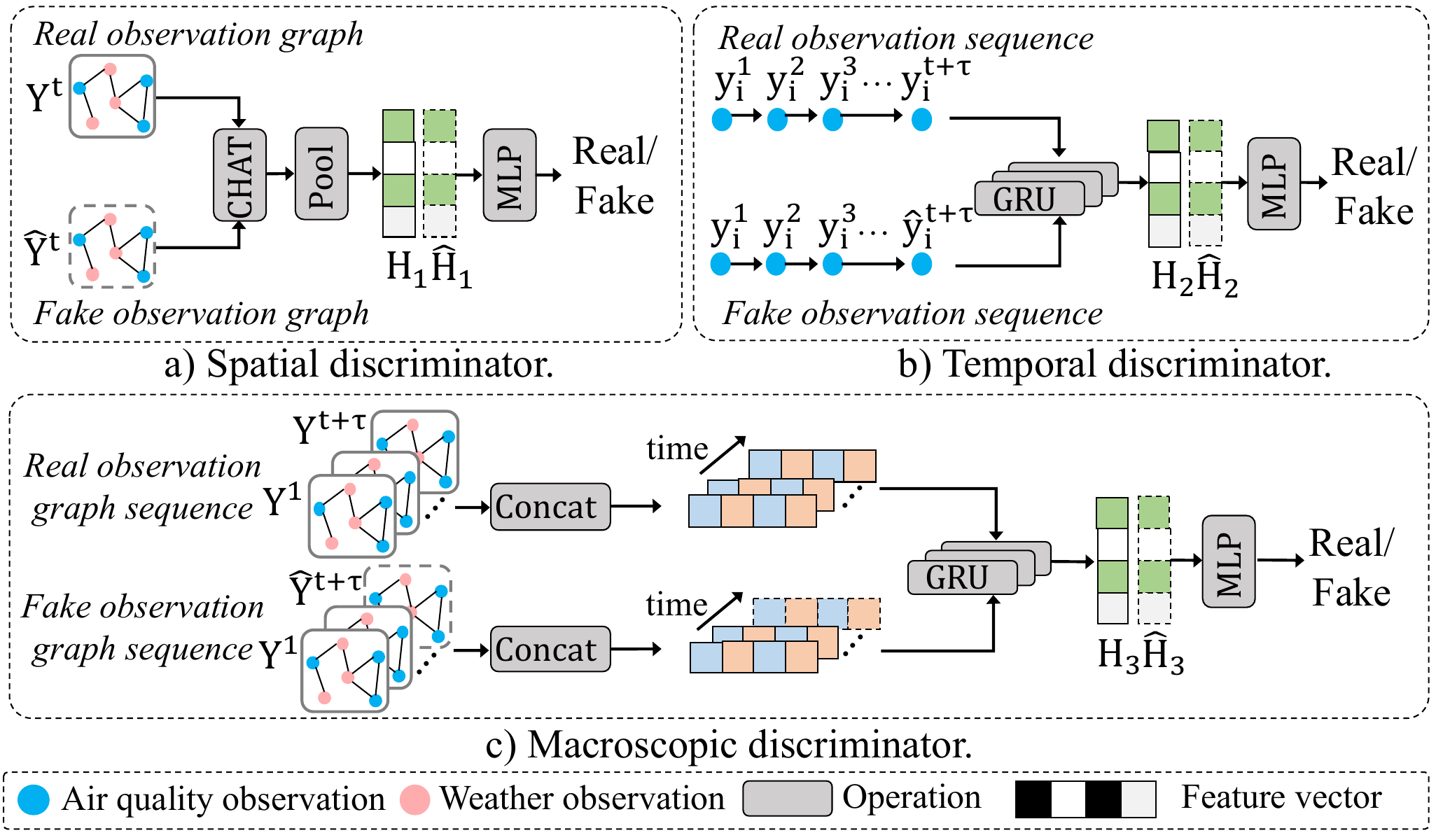}
  \vspace{-0.5cm}
  \caption{Discriminators in multi-adversarial learning.}\label{adver_train}
  \vspace{-0.7cm}
\end{figure}

We further introduce the multi-adversarial learning framework to obtain robust station representations.
By proactively simulating and resisting noisy observations via a \emph{minmax} game between the generator and discriminator~\cite{goodfellow2014generative}, adversarial learning can make the generator more robust and generalize better to the underlying true distribution $p_{true}(\mathbf{Y} | \mathbfcal{X}, \mathbf{C})$.
%resist observation noise propagation, including one generator~(\ie the prediction model) and multiple discriminators. \TODO{More insights.}
Compared with standard adversarial learning, our multi-adversarial learning framework encourages the generator to make more consistent predictions from both spatial and temporal domains.
Specifically, we function the HRGNN $\mathbf{\phi}$ as the generator $\hat{\mathbf{Y}}=G(\mathbfcal{X}, \mathbf{C};\mathbf{\phi})$ for joint predictions.
Moreover, we propose a set of distinct discriminators $\{D_k(\cdot;\theta_k)\}^K_{k=1}$ to distinguish the real observations and predictions from both microscopic and macroscopic perspectives, as detailed below.

\subsubsection{Microscopic discriminators.}
Microscopic discriminators aims at enforcing the generator to approximate the underlying spatial and temporal distributions, \ie pair-wise spatial autocorrelation and stepwise temporal autocorrelation.
%To this end, we further construct microscopic discriminators from spatial and temporal domains.

\emph{Spatial discriminator}. From the spatial domain, the adversarial perturbation induces the compounding propagation error through the HSG, as illustrated in Figure~\ref{adver_train}~(a). Consider a time step $t$, the spatial discriminator $D_s(\mathbf{y}^t;\theta_s)$ aims to maximize the accuracy of distinguishing city-wide real observations and predictions at the current time step,
\begin{equation}\label{equ:dis1}
\mathcal{L}_{s}=\log D_s(\mathbf{y}^t;\theta_s)+log(1-D_s(\hat{\mathbf{y}}^t;\theta_s)),
\end{equation}
where $D_s(\mathbf{y}^t;\theta_s)$ is parameterized by the context-aware heterogeneous graph attention block in HRGNN followed by a multi-layer perception, $\mathbf{y}^t$ and $\hat{\mathbf{y}}^t$ are ground truth observations and predicted conditions of the city, respectively.

\emph{Temporal discriminator.} From the temporal domain, the adversarial perturbation induces accumulated error for each station from previous time steps. 
%Similar to the spatial discriminator, 
As depicted in Figure~\ref{adver_train}~(b), the temporal discriminator $D_t(\mathbf{y}_i;\theta_t)$ outputs a probability indicating how likely a observation sequence $\mathbf{y}_i$ of a particular station $s_i$ is from the real data distribution $p_{true}(\mathbf{y}_i | \mathbf{x}_i, \mathbf{c}_i)$ rather than the generator $p_{\phi}(\mathbf{y}_i | \mathbf{x}_i, \mathbf{c}_i)$.
%classifies ground truth observation $\mathbf{y}_i$ and prediction $\hat{\mathbf{y}}_i$ of a air quality or weather station $s_i$ in past $T$ and future $\tau$ time steps, as depicted in \TODO{Figure 2 (b)}. 
\begin{equation}\label{equ:dis2}
\mathcal{L}_{t}=\log D_t(\mathbf{y}_i;\theta_t)+log(1-D_t(\hat{\mathbf{y}}_i;\theta_t)).
\end{equation}
Different from the spatial discriminator, $D_t(\mathbf{y}_i;\theta_t)$ is parameterized by the temporal block in HRGNN followed by a multi-layer perception. 
$\mathbf{y}_i$ is a sequence of observations in past $T$ and future $\tau$ time steps.
%$D_t(\mathbf{y}_i;\theta_t)$ outputs a probability indicating how likely $\mathbf{y}_i$ is from the real data distribution $p_{true}$ rather than the generator $p_{\phi}$.
%The temporal discriminator aims to maximize 

\subsubsection{Macroscopic discriminator.}
As illustrated in Figure~\ref{adver_train}~(c), we further propose a macroscopic discriminator to capture the globally underlying distribution, denoted by $D_m(\mathbf{Y};\theta_m)$. In particular, the macroscopic discriminator aims to maximize the accuracy of distinguishing the ground truth observations and predictions generated from $G$ from a global view,
\begin{equation}\label{equ:dis3}
\mathcal{L}_{m}=\log D_m(\mathbf{Y};\theta_m)+log(1-D_m(G(\mathbfcal{X},\mathbf{C};\mathbf{\phi});\theta_m)),
\end{equation}
where $D_m(\mathbf{Y};\theta_m)$ is parameterized by a GRU followed by a multi-layer perception.
Note that for efficiency concern, the input of $D_m(\mathbf{Y};\theta_m)$ is a simple concatenation of observations from all stations $S$ in past $T$ and future $\tau$ time steps, but other  graph convolution operations are also applicable.

\eat{
Given the generator $G$ and discriminators $\{D_i\}^K_{i=1}$,
the minmax function of the multi-adversarial learning is
\begin{equation}
\begin{aligned}
\min_{G} \max_{\{D_i\}^K_{i=1}} V(\{D_i\}^K_{i=1}&,G)=\mathbb{E}_{\mathbf{Y} \sim p_{\emph{true}}}[\log D_i(\mathbf{Y};\theta_i)]+\\
&\mathbb{E}_{\mathbfcal{X} \sim p_{\phi}}[\log (1-D_i(G(\mathbfcal{X},\mathbf{C};\phi)))].
%\min_{G} \max_{\{D_i\}^K_{i=1}} V(D_i,G)=&\mathbb{E}_{\mathbf{Y} \sim p_{\emph{true}}(\mathbf{Y}|\mathbfcal{X},\mathbf{C})}[\log D_i(\mathbf{Y})]\\
%&+\mathbb{E}_{\hat{\mathbf{Y}} \sim p_{\phi}(\hat{\mathbf{Y}}|\mathbfcal{X},\mathbf{C})}[\log (1-D_i(G(\hat{\mathbf{Y}})))],
\end{aligned}
\end{equation}
}

\subsection{Multi-Task Adaptive Training}
It is widely recognized that adversarial training suffers from the unstable and mode collapse problem, where the generator is easy to over-optimized for a particular discriminator~\cite{salimans2016improved,neyshabur2017stabilizing}.
To stabilize the multi-adversarial learning and achieve overall accuracy improvement, we introduce a multi-task adaptive training strategy to dynamically re-weight discriminative loss and enforce the generator to perform well in various spatiotemporal aspects.
%Consider the ground truth observation $\mathbfcal{Y}$ and prediction $\hat{\mathbfcal{Y}}$, we first measure the 

%In summary, the loss function of MasterGNN includes (1) minimizing the predictive loss $\mathcal{L}_g$~(Equation~\ref{equ:mse}), and (2) maximizing discriminative losses $\mathcal{L}_D=\{\mathcal{L}_{d_i}\}^K_{i=1}$~(Equation~\ref{equ:dis1}-\ref{equ:dis3}). 
Specifically, the objective of MasterGNN is to minimize 
\begin{equation}\label{equ:all-loss}
\mathcal{L}=\mathcal{L}_g+\sum^K_{i=1}\lambda_i \mathcal{L}_{d_i},
\end{equation}
where $\mathcal{L}_g$~(Equation~\ref{equ:mse}) is the predictive loss, $\mathcal{L}_D=\{\mathcal{L}_{d_i}\}^K_{i=1}$ (Equation~\ref{equ:dis1}-\ref{equ:dis3}) are discriminative losses, and
$\lambda_i$ is the importance of the corresponding discriminative loss.

Suppose $\mathbf{H}_{i}$ and $\hat{\mathbf{H}}_{i}$ are intermediate hidden states in discriminator $D_i$ based on real observations $\mathbf{Y}$ and predictions $\hat{\mathbf{Y}}$.
We measure the divergence between $\mathbf{H}_{i}$ and $\hat{\mathbf{H}}_{i}$ by $\gamma_i=sim(\sigma(\mathbf{H}_{i}),\sigma(\hat{\mathbf{H}}_{i}))$,
where $sim(\cdot,\cdot)$ is an Euclidean distance based similarity function, $\sigma$ is the sigmoid function. 
Intuitively, $\gamma_i$ reflects the hardness of $D_i$ to distinguish the real sample.
In each iteration, we re-weight discriminative losses by 
%\begin{equation}
$\lambda_i = \frac{exp(\gamma_i)}{\sum^{K}_{k=1}exp(\gamma_k)}$.
%\end{equation}
In this way, the generator pays more attention to discriminators with a larger room to improve, and will result in better prediction performance.

\section{Experiments}

\begin{table}[t]
\centering
\begin{small}
%\vspace{-2pt}
\caption{Statistics of datasets.}
\label{table-data-stats}
\vspace{-6pt}
\begin{tabular}{c | c  c} \hline
{Data description} & Beijing & Shanghai \\ \hline \hline
\# of air quality stations & 35 & 76 \\ 
\hline
\# of weather stations & 18 & 11\\ 
\hline
\# of air quality observations & 409,853 & 1,331,520 \\ 
\hline
\# of weather observations & 210,824 & 192,720\\ 
\hline
%Time span & 1/1/2017/4/1/2018 & 1/1/2018-4/1/2018\\
%\hline
\# of  POIs & 900,669 & 1,061,399\\ 
\hline
\# of road segments & 812,195 & 768,336 \\ 
\hline
\end{tabular}
\end{small}
\vspace{-7mm}
\end{table}

\begin{table*}[t]
\centering
\vspace{-10pt}
\begin{small}
\caption{Overall performance comparison of different approaches. A smaller value indicates a better performance.}
\vspace{-5pt}
\label{tab-eff}
\begin{tabular}{c|c|c|c|c|c|c|c|c}%
\hline
\multirow{3}{*}{Model} & \multicolumn{8}{|c}{MAE/SMAPE}\\
\cline{2-9}
\multirow{3}{*}{} & \multicolumn{4}{|c}{Beijing} & \multicolumn{4}{|c}{Shanghai}\\
\cline{2-9}
\multirow{3}{*}{} & {AQI} & {Temperature} & {Humidity} & {Wind speed} & {AQI} & {Temperature} & {Humidity} & {Wind speed}\\
\hline
ARIMA & 54.97/0.925 & 3.26/0.475 & 16.42/0.403 & 1.17/0.637 & 34.36/0.597 & 2.27/0.394 & 11.96/0.267 & 1.35/0.653 \\
%\hline
LR  & 45.27/0.732 & 2.39/0.426 & 9.58/0.296 & 1.04/0.543 & 27.68/0.462 & 1.93/0.342 & 9.53/0.254 & 1.21/0.568 \\
%\hline
GBRT  & 38.36/0.678 &  2.31/0.397 & 8.43/0.254 & 0.93/0.489 & 24.13/0.423 & 1.87/0.329 & 8.91/0.232 & 1.05/0.497 \\
\hline
%\hline
DeepAir & 31.45/0.613 & 2.21/0.375 & 8.25/0.246 & 0.82/0.468 & 18.79/0.297 & 1.79/0.304 & 7.82/0.209 & 0.93/0.516 \\
%\hline
GeoMAN & 30.04/0.595 & 2.05/0.356 & 8.01/0.223 & 0.71/0.457 & 20.75/0.348 & 1.56/0.261 & 6.45/0.153 & 0.82/0.465 \\
%\hline
GC-DCRNN & 29.58/0.586 & 2.14/0.364 & 8.14/0.235 & 0.73/0.462 & 18.73/0.305 & 1.61/0.263 & 6.83/0.186 & 0.78/0.462 \\
%\hline
DUQ & 32.69/0.624 & 2.02/0.348 & 7.97/0.217 & 0.69/0.446 & 22.81/0.364 & 1.42/0.235 & 6.23/0.136 & 0.67/0.442 \\
\hline\hline
MasterGNN & \textbf{27.45/0.548} & \textbf{1.87/0.326} & \textbf{7.25/0.184}& \textbf{0.64/0.428} &\textbf{16.51/0.265} & \textbf{1.25/0.213} & \textbf{5.64/0.095}&\textbf{0.61/0.427}\\
\hline
\end{tabular}
\vspace{-8pt}
\end{small}
\end{table*}

\subsection{Experimental settings}
\subsubsection{Data description.}
We conduct experiments on two real-world datasets collected from Beijing and Shanghai, two metropolises in China. 
%We use two real-world datasets collected from Beijing and Shanghai, two metropolises in China, for evaluation.
The Beijing dataset is ranged from January 1, 2017, to April 1, 2018, and the Shanghai dataset is ranged from June 1, 2018, to June 1, 2020.
Both datasets include (1) air quality observations~(\ie PM2.5, PM10, O3, NO2, SO2, and CO), (2) weather observations and weather forecast~(\ie weather condition, temperature, pressure, humidity, wind speed and wind direction), and (3) urban contextual data~(\ie POI distribution and Road network distribution~\cite{zhu2016days,liu2019hydra,liu2020polestar}). 
%In each dataset, the air quality observations include PM2.5, PM10, O3, NO2, SO2, and CO, the weather observations include weather condition, temperature, pressure, himidity, wind speed and wind direction, and the urban contextual data include POI distribution and Road network distribution~\cite{liu2019hydra}.
The observation data of Beijing dataset is from KDD CUP 2018\footnote{https://www.biendata.xyz/competition/kdd\_2018/data/}, and the observation data of Shanghai dataset is crawled from China government websites\footnote{http://www.cnemc.cn/en/}.
All air quality and weather observations are collected by every hour. 
We associate POI and road network features to each station through an open API provided by an anonymized online map application.
Same as existing studies~\cite{yi2018deep,wang2019deep}, we focus on Air Quality Index~(AQI) for air quality prediction, which is derived by the Chinese AQI standard, and temperature, humidity and wind speed for weather prediction.
We split the data into training, validation, and testing data by the ratio of 8:1:1.
The statistics of two datasets are shown in Table~\ref{tab-eff}.
%For the Beijing dataset, the raw air quality data is from he KDD CUP 2018\footnote{https://www.biendata.xyz/competition/kdd\_2018/data/}, and we complement the 
%The raw data of Beijing is a public dataset ranged from January 1, 2017, to April 1, 2018, that comes from the KDD CUP 2018\footnote{https://www.biendata.xyz/competition/kdd\_2018/data/}. The second dataset covers air quality and  weather stations in Shanghai for over two years\footnote{http://data.epmap.org/} (from June 1, 2018, to June 1, 2020). 
%According to the Chinese AQI standard, air pollution concentration is converted into the corresponding AQI. 
%In addition, the context data are collected through a large navigation app, we associate POI distribution and road network density to each station. 

\subsubsection{Implementation details.}
Our model and all the deep learning baselines are implemented with PyTorch. All methods are evaluated on a Linux server with 8 NVIDIA Tesla P40 GPUs.
%PaddlePaddle\footnote{https://github.com/PaddlePaddle}. 
We set context-aware heterogeneous graph attention layers $l=2$.
%for spatial autocorrelation modeling. 
The cell size of the GRU is set to $64$.  %During the training phase, the hyper-parameters $\epsilon$, $\lambda$ are set to 5, 0.05, respectively. 
We set $\epsilon=15$, and the hidden size of the multi-layer perceptron of each discriminator is fixed to $64$. All the learning parameters are initialized with a uniform distribution, and the model is trained by stochastic gradient descent with a learning rate $lr=0.00001$. The activation function used in the hidden layers is LeakyReLU ($\alpha$=0.2).
All the baselines use the same input features as ours, except ARIMA excludes contextual features. 
Each numerical feature is normalized by Z-score. We set $T=72$ and the future time step $\tau=48$.
For a fair comparison, we carefully fine-tuned the hyper-parameters of each baseline, and use one another historical observations as a part of feature input.
%Please refer to the supplementary material and source code for more details.

%Our model is evaluated on a powerful Linux server with 8 NVIDIA Tesla P40 GPUs.

\subsubsection{Evaluation metrics.}
We use Mean Absolute Error (MAE) and Symmetric Mean Absolute Percentage Error (SMAPE), two widely used metrics~\cite{luo2019accuair} for evaluation.

\subsubsection{Baselines.}
We compare the performance of MasterGNN with the following seven baselines:

%and two variants of our model:
% \noindent\textbf{ARIMA} ~\cite{brockwell2002introduction} is a classic time series prediction method utilizes the moving average and auto-regressive component to estimate future trends.
% \textbf{LR} uses linear regression ~\cite{montgomery2012introduction} for joint prediction. We concatenate previous observations and contextual features as input.
% \textbf{GBRT} is a tree-based model widely used for regression tasks. We adopt an efficient version XGBoost~\cite{chen2016xgboost}, and the input feature is same as LR. 
% \textbf{GeoMAN}~\cite{liang2018geoman} is a multi-level attention-based network which integrates local spatial attention, global spatial attention, and temporal attention for geo-sensory time series prediction.
% \textbf{DeepAir}~\cite{yi2018deep} is a deep learning method for air quality prediction. It combines spatial transformation and distributed fusion components to fuse various urban data.
% \textbf{GC-DCRNN}~\cite{lin2018exploiting} leverages a diffusion convolution recurrent neural network to model spatiotemporal dependencies for air quality prediction. We follow the graph construction method described in the original paper.
% \textbf{DUQ}~\cite{wang2019deep} is an advanced weather forecasting framework that improves numerical weather prediction by using uncertainty quantification and a dedicated negative log-likelihood error loss function.
\begin{itemize}
    \item \textbf{ARIMA} ~\cite{brockwell2002introduction} is a classic time series prediction method utilizes the moving average and auto-regressive component to estimate future trends.
    \item \textbf{LR} uses linear regression ~\cite{montgomery2012introduction} for joint prediction. We concatenate previous observations and contextual features as input.
    %extract hand-crafted features from previous historical data as model input.
    \item \textbf{GBRT} is a tree-based model widely used for regression tasks. We adopt an efficient version XGBoost~\cite{chen2016xgboost}, and the input feature is same as LR.
    \item \textbf{GeoMAN}~\cite{liang2018geoman} is a multi-level attention-based network which integrates spatial and temporal attention for geo-sensory time series prediction.
    \item \textbf{DeepAir}~\cite{yi2018deep} is a deep learning method for air quality prediction. It combines spatial transformation and distributed fusion components to fuse various urban data.
    \item \textbf{GC-DCRNN}~\cite{lin2018exploiting} leverages a diffusion convolution recurrent neural network to model spatiotemporal dependencies for air quality prediction.
    \item \textbf{DUQ}~\cite{wang2019duq} is an advanced weather forecasting framework that improves numerical weather prediction by using uncertainty quantification.
\end{itemize}

\begin{figure*}[t]
	\begin{minipage}{1.\linewidth}
		\centering
		\vspace{0ex}
		%\hspace{2ex}
		\subfigure[{\small Effect of joint prediction.}]{\label{copred}
			\includegraphics[width=0.24\textwidth]{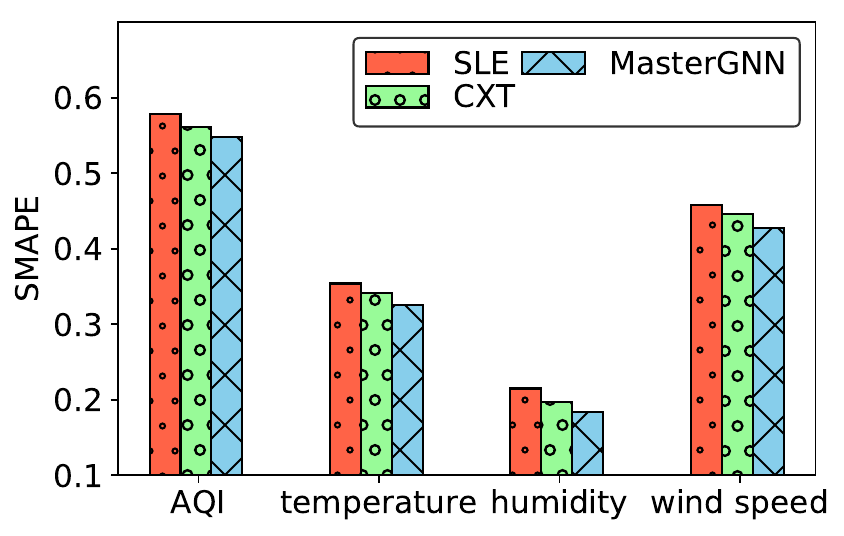}}
		\subfigure[{\small Effect of heterogeneous recurrent graph neural network.}]{\label{heterst}
			\includegraphics[width=0.24\textwidth]{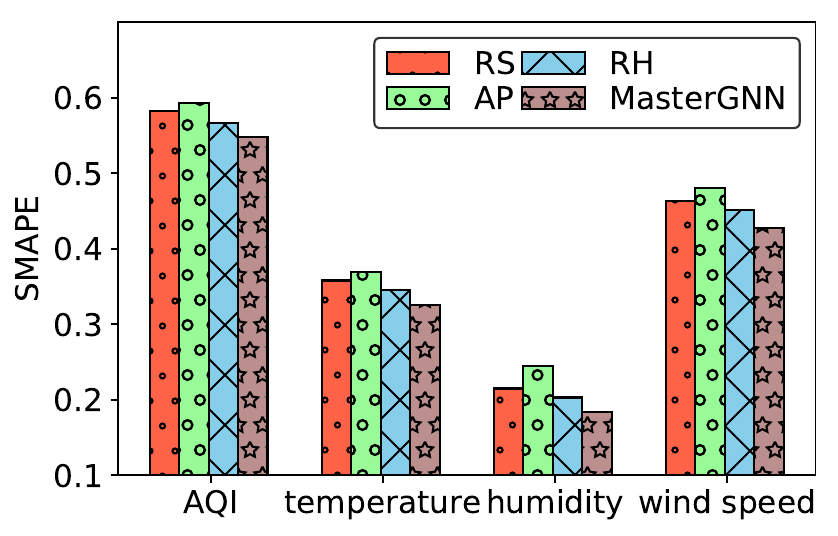}}
		\subfigure[{\small Effect of multi-adversarial learning.}]{\label{adv}
			\includegraphics[width=0.24\textwidth]{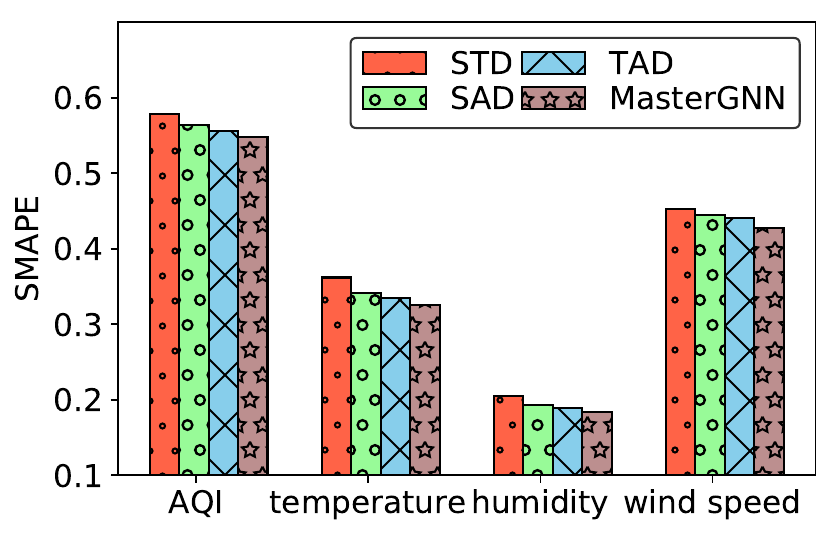}}
		\subfigure[{\small Effect of multi-task adaptive training.}]{\label{opti}
			\includegraphics[width=0.24\textwidth]{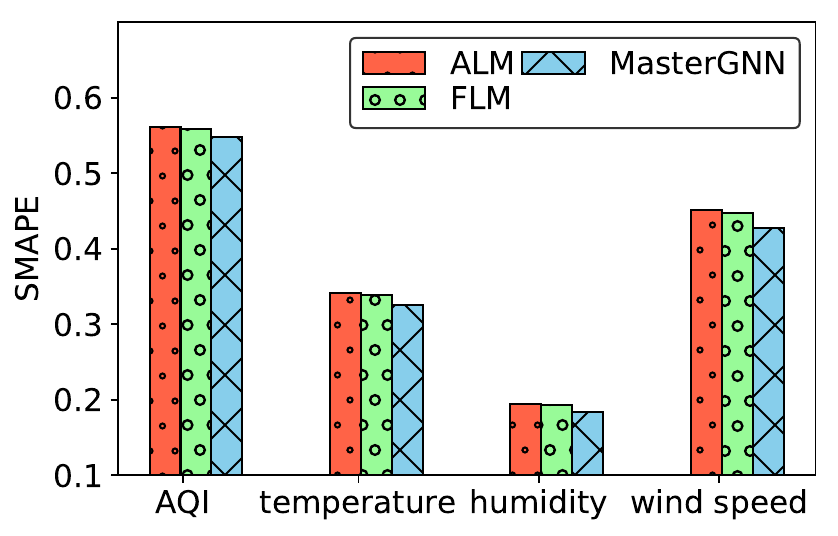}}
	\end{minipage}
	\vspace{-3ex}
	\caption{Ablation study of MasterGNN on the Beijing dataset.} 
	\vspace{-7mm}
	\label{parasensitivity}
\end{figure*}

\subsection{Overall performance}
Table 2 reports the overall performance of MasterGNN and compared baselines on two datasets with respect to MAE and SMAPE. 
%Note that all the baselines predict air quality and meteorology separately. 
We observe our approach consistently outperforms all the baselines using both metrics, which shows the superiority of MasterGNN.
%It is obvious that our approach outperforms all the baselines using both metrics, which clearly proves the advance of MasterGNN.
Specifically, our model achieves (7.8\%, 8.0\%, 10.0\%, 7.8\%) and (6.9\%, 6.7\%, 17.9\%, 4.2\%) improvements beyond the best baseline on MAE and SMAPE on Beijing for (AQI, temperature, humidity, wind speed) prediction, respectively. Similarly, the improvement of MAE and SMAPE on Shanghai are (13.4\%, 13.6\%, 10.4\%, 9.8\%) and (12.1\%, 10.3\%, 43.2\%, 3.5\%), respectively. 
Moreover, we observe all deep learning based models outperform statistical learning based approaches by a large margin, which demonstrate the effectiveness of deep neural network for modeling spatiotemporal dependencies. 
Remarkably, GC-DCRNN performs better than all other baselines for air quality prediction, and DUQ consistently outperforms other baselines for weather prediction. 
However, they perform relatively poorly on the other task, which indicates their poor generalization capability and further demonstrate the advantage of joint prediction.
%Additionally, GBRT is the best statistical learning method, which validates our expectation that GBRT can capture sophisticated non-linear air quality and weather patterns for prediction.
%The main reason is that GC-DCRNN can better model spatial correlation and diffusion process of air pollutants. 
%For weather forecasting task, DUQ consistently outperform other baselines, which show the advantage of uncertainty quantification
%Although GC-DCRNN and DUQ are the respectively best baseline for air quality and weather prediction, they perform relatively poor on the other task, which indicates their poor generalization capability.
%ARIMA and LR have the worst performance due to its incapability of capturing complex spatiotemporal dependencies. 
%GBRT in handling regression tasks.

\begin{figure*}[t]
	\begin{minipage}{1.\linewidth}
		\centering
		\vspace{-10pt}
		%\hspace{2ex}
		\subfigure[{\small Effect of $T$.}]{\label{length}
			\includegraphics[width=0.24\textwidth]{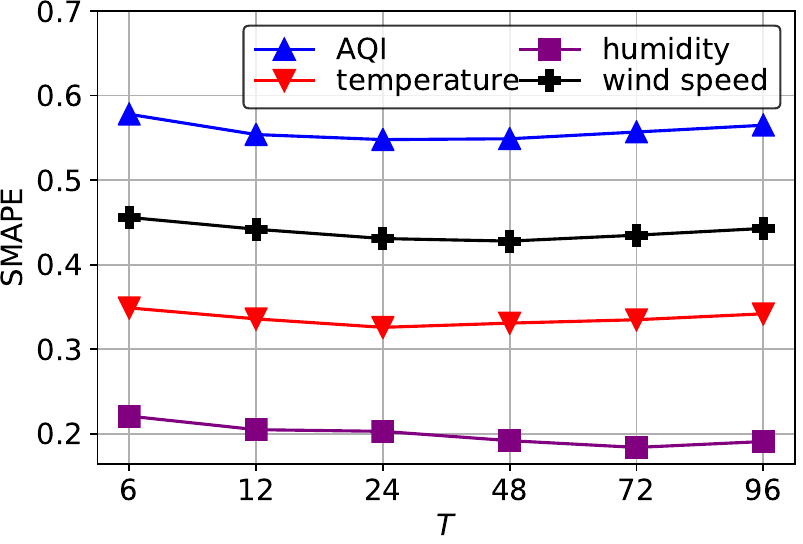}}
		\subfigure[{\small Effect of $d$.}]{\label{hidden_units}
			\includegraphics[width=0.24\textwidth]{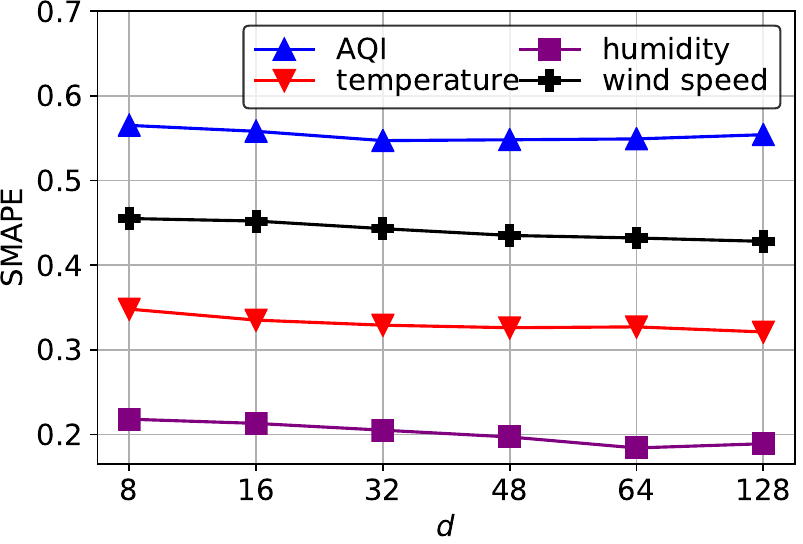}}
		\subfigure[{\small Effect of $\epsilon$.}]{\label{epsilon}
			\includegraphics[width=0.24\textwidth]{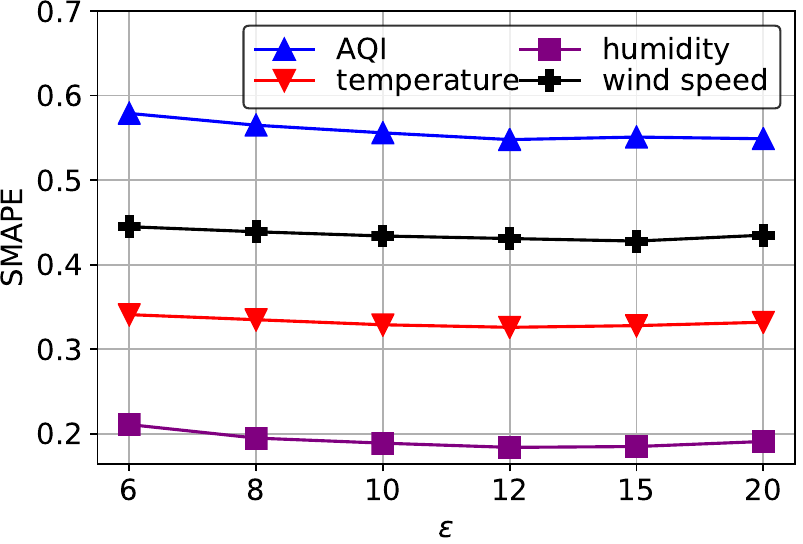}}
		\subfigure[{\small Effect of $\tau$.}]{\label{tau}
			\includegraphics[width=0.24\textwidth]{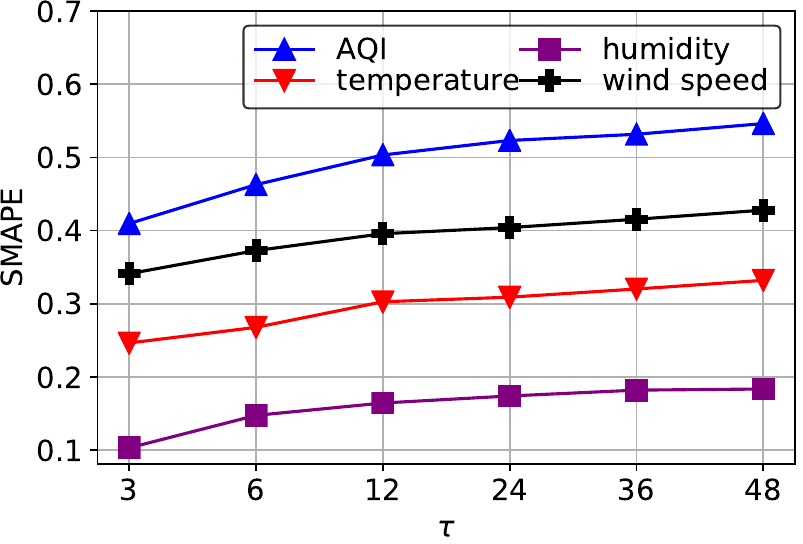}}
	\end{minipage}
	\vspace{-3ex}
	\caption{Parameter sensitivities on the Beijing dataset.} 
	\vspace{-3ex}
	\label{parasensitivity}
\end{figure*}

\subsection{Ablation study}
Then we study the effectiveness of each module in MasterGNN. Due to the page limit, we only report the results on Beijing by using SMAPE. The results on Beijing using MAE and on Shanghai using both metrics are similar.

\textbf{Effect of joint prediction}. To validate the effectiveness of joint prediction, we compare the following variants of MasterGNN:
(1) SLE models two tasks independently without information sharing,
%without joint prediction and the other observation information,
%only uses the observation we aim to predict as input, 
(2) CXT models two tasks independently but uses the other observations as a part of contextual features. 
%without joint prediction but use the other observations as a part of contextual features. 
As depicted in Figure \ref{copred}, we observe a performance gain of CXT by adding the other observation features, but MasterGNN achieves the best performance.
Such results demonstrate the advantage of modeling the interactions between two tasks compared with simply adding the other observations as the feature input.

\textbf{Effect of heterogeneous recurrent graph neural network}. We evaluate three variants of HRGNN, (1) RS removes the spatial autocorrelation block from MasterGNN, (2) AP replaces GRU by average pooling, (3) RH handles air quality and weather observations homogeneously. As shown in Figure \ref{heterst}, we observe MasterGNN achieves the best performance compared with other variants, demonstrating the benefits of incorporating spatial and temporal autocorrelations for joint air quality and weather prediction. 
%In particular, MasterGNN achieves a better performance than GAT, this observation indicates its advantage in capturing spatial interactions among heterogeneous stations.

\textbf{Effect of multi-adversarial learning}.
We compare the following variants: (1) STD removes the macro discriminator, (2) SAD removes the micro temporal discriminator, (3) TAD removes the micro spatial discriminators. As shown in Figure \ref{adv}, the macro discriminator is the most influential one, which indicates the importance of defending noises from the global view.
Overall, MasterGNN consistently outperforms STD, SAD, and TAD on all tasks by integrating all discriminators, demonstrating the benefits of developing multiple discriminators for joint prediction. 
%Besides, the SAD and STD performs better than STD, which demonstrates the proposed approach can obtain more benefits using macro discriminator.

\textbf{Effect of multi-task adaptive training}.
To verify the effect of multi-task adaptive training, we develop (1) Average Loss Minimization (ALM) uses the same weight for all discriminators, (2) Fixed Loss Minimization~(FLM) sets weighted but fixed importance based on the performance loss of MasterGNN when the corresponding discriminator is removed.
%instead of the multi-task adaptive training and \TODO{fixed importance optimization}. 
As shown in Figure \ref{opti}, we observe performance degradation when we fix each discriminator's weight, either on average or weighted. 
Overall, adaptively re-weighting each discriminator's importance can guide the optimization direction of the generator and lead to better performance. 
%Our training strategy can adaptively guide the generator performs well in various spatiotemporal aspects and achieves the purpose of balancing discriminative losses. 

\subsection{Parameter sensitivity}

We further study the parameter sensitivity of MasterGNN.
%including the impact of the input length $T$, hidden state dimension $d$,  distance threshold $\epsilon$ and prediction steps $\tau$. 
We report SMAPE on the Beijing dataset. Each time we vary a parameter, we set others to their default values.
%Due to the space limitations, we report the impact of the input length $T$, hidden state dimension $d$,  distance threshold $\epsilon$ and prediction horizon $\tau$ using SMAPE on Beijing. Each time we vary a parameter, we set others to their default values.

First, we vary the input length $T$ from 6 to 96. The results are reported in Figure ~\ref{length}. As the input length increases, the performance first increases and then gradually decreases. The main reason is that a short sequence cannot provide sufficient temporal periodicity and trend information. But too large input length may introduce noises for future prediction, leading to performance degradation.

Then, we vary  $d$ from 8 to 128. The results are reported in Figure ~\ref{hidden_units}. We can observe that the performance first increases and then remains stable. However, too large $d$ leads to extra computation overhead. Therefore, set the $d=64$ is enough to achieve a satisfactory result.

After that, to test the impact of the distance threshold in HSG, we vary $\epsilon$ from 6 to 20. The results are reported in Figure ~\ref{epsilon}. %Increasing $\epsilon$ encourage the model to incorporate the spatial autocorrelation between more monitoring stations. 
As $\epsilon$ increases, we observe a performance first increase then slightly decrease, which is perhaps because a large $\epsilon$ integrates too many stations, which introduces more noises during spatial autocorrelation modeling.

Finally, we vary the prediction step $\tau$ from 3 to 48. The results are reported in Figure ~\ref{tau}. We observe the performance drops rapidly when $\tau$ goes large. The reason perhaps is the uncertainty increases when $\tau$ is large, and it is difficult for the machine learning model to quantify such uncertainty.

\subsection{Qualitative study}
\begin{figure}
  \centering
  \vspace{-4pt}
  \includegraphics[width=8cm]{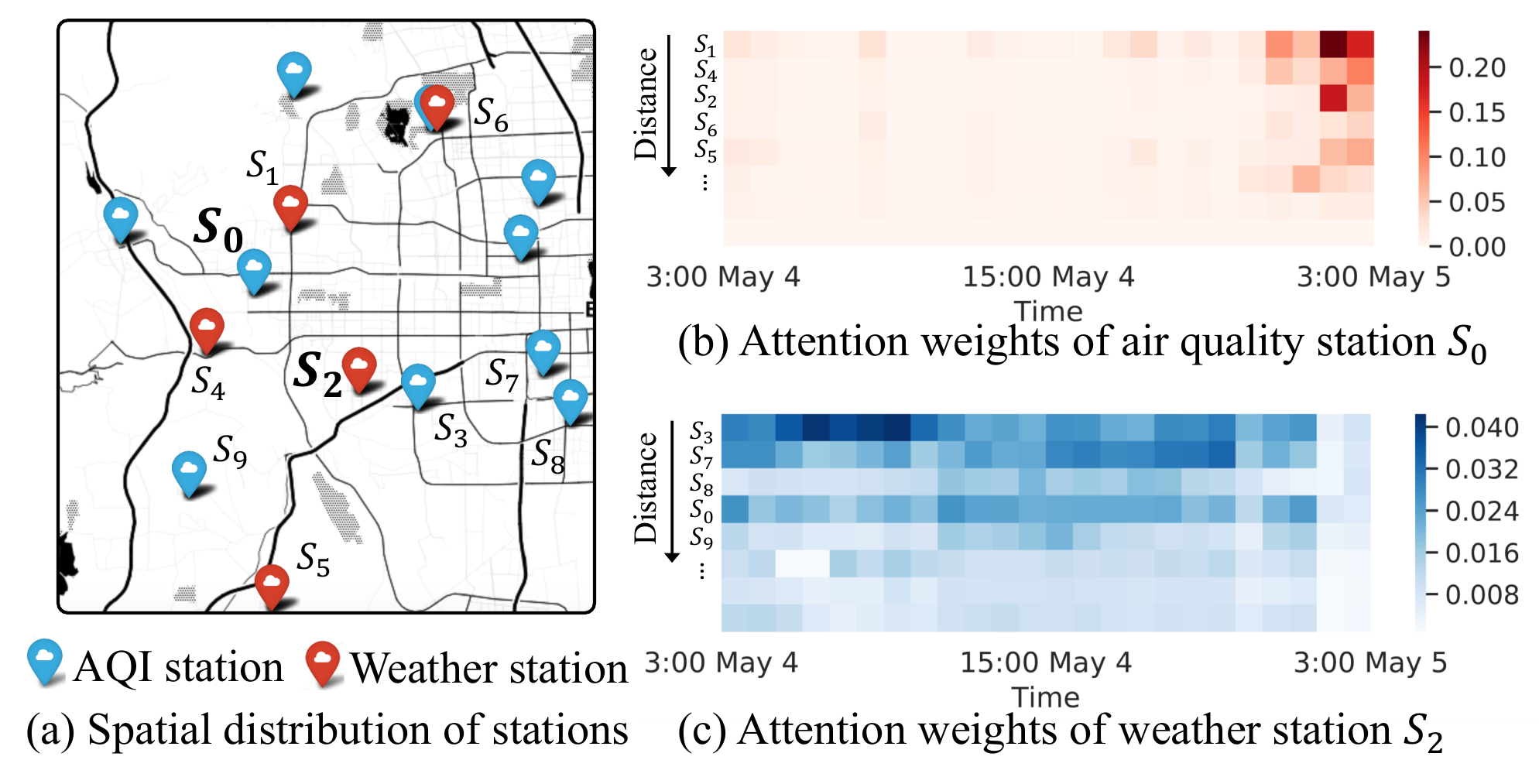}
  \vspace{-4pt}
  \caption{Visualization of the learned attention weight of neighboring air quality and weather stations.}\label{fig:casestudy}
  \vspace{-7mm}
\end{figure}

Finally, we qualitatively analyze why MasterGNN can yield better performance on both tasks. 
Figure~\ref{fig:casestudy}~(a) shows the distribution of air quality and weather monitoring stations in the Haidian district, Beijing. 
We perform a case study from 3:00 May 4, 2017, to 3:00 May 5, 2017.
First, we take the air quality station $S_0$ as an illustrative example to show how weather stations help air quality prediction.
In MasterGNN, the attention score reflects the relative importance of each neighboring station for prediction.
Figure~\ref{fig:casestudy}~(b) visualizes the attention weights of eight neighboring weather stations of $S_0$.
As can be seen, the attention weights of weather stations $S_1$, $S_2$ and $S_4$ abruptly increase because a strong wind blows through this area, which results in notable air pollutant dispersion in the next five hours.
Clearly, the weather stations provide additional knowledge for air quality prediction.
Then, we take the weather station $S_2$ to depict air quality stations' impact on weather prediction.
Figure~\ref{fig:casestudy}~(c) shows the attention weights of neighbouring station of $S_2$.
We observe the nearest air quality station $S_3$ is the most influential station during 5:00 May 4 2017 and 10:00 May 4 2017, while $S_7$ plays a more important role during 15:00 May 4 2017 and 21:00 May 4 2017, corresponding to notable air pollution concentration changes during these periods.
The above observations further validate the additional knowledge introduced by the joint modeling of heterogeneous stations for both predictive tasks.

\section{Related work}

\textbf{Air quality and weather prediction}.
%Air quality and weather forecasting have been investigated for many years. 
%Existing studies on air quality and weather prediction can be divided into numerical-based methods and learning-based methods. Numerical methods simulate the dispersion process of atmospheric elements based on current observations and empirical assumptions~\cite{liu2005prediction,lorenc1986analysis}, which lead to inaccurate and inefficient prediction. 
%For learning based methods, statistical models such as ARIMA \cite{chen2011comparison} and SVM~\cite{wang2014research} was first applied to capture data correlations.
Existing literature on air quality and weather prediction can be categorized into two classes.
(1) \emph{Numerical-based models} make predictions by simulating the dispersion of various air quality or weather elements based on physical laws~\cite{lorenc1986analysis,vardoulakis2003modelling,liu2005prediction,richardson2007weather}.
(2) \emph{Learning-based models} utilize end-to-end machine learning methods to capture spatiotemporal correlations based on historical observations and various urban contextual data~(\eg POI distributions, traffics)~\cite{chen2011comparison,zheng2013u,cheng2018neural}.
Recently, many deep learning models~\cite{yi2018deep,wang2019deep,lin2018exploiting} have been proposed to enhance the performance of air quality and weather prediction.
By leveraging the representation capacity of deep learning for spatiotemporal autocorrelation modeling, learning-based models usually achieve better prediction performance than numerical-based models.
Unlike the above approaches, our method explicitly models the correlations and interactions between the air quality and weather prediction task and achieves a better performance.
%For example, DeepAir~\cite{yi2018deep} learning and fusing embeddings of various urban factors for air quality prediction. DUQ~\cite{wang2019deep} integrates a uncertainty quantification method with the sequence-to-sequence network for weather forecasting.
%Moreover, the development of graph neural network has also promoted the accurate air quality prediction. For example, GC-DCRNN~\cite{lin2018exploiting} employs a geo-context based diffusion convolutional recurrent neural network for air quality prediction.
%Our approach is distinguished from these methods in three aspects.
%via a heterogeneous recurrent graph neural network.

\textbf{Adversarial Learning}.
Adversarial learning~\cite{goodfellow2014generative} is an emerging learning paradigm for better capturing the data distribution via a minmax game between a generator and a discriminator.
%Classical generative adversarial networks consist of two components: a generator tries to generate fake samples using random noise vectors, and a discriminator tries to distinguish whether the sample came from the generator or real dataset. 
%Many works have shown that GAN can be successfully applied to real-world applications, such as image translation~\cite{isola2017image}, super resolution~\cite{ledig2017photo} and time series imputation~\cite{luo2018multivariate}.
In the past years, adversarial learning has been widely applied to many real-world application domains, such as sequential recommendation~\cite{yu2017seqgan} and graph learning~\cite{Wang2018GraphGANGR}.
%However, it is widely recognized that GAN is hard to optimize due to the instability of training and mode collapse problem~\cite{salimans2016improved}. 
Recently, we notice several multi-adversarial frameworks have been proposed for improving image generation tasks~\cite{nguyen2017dual,hoang2018mgan,albuquerque2019multi}.
Inspired by the above studies, we extend the multi-adversarial learning paradigm to the environmental science domain and introduce an adaptive training strategy to improve the stability of adversarial learning.

\section{Conclusion}
In this paper, we presented MasterGNN, a joint air quality and weather prediction model that explicitly models the correlations and interactions between two predictive tasks.
%based on historical atmospheric observations and urban contextual factors. 
Specifically, we first proposed a heterogeneous recurrent graph neural network to capture spatiotemporal autocorrelation among air quality and weather monitoring stations. Then, we developed a multi-adversarial learning framework to resist observation noise propagation. In addition, an adaptive training strategy is devised to automatically balance the optimization of multiple discriminative losses in multi-adversarial learning. Extensive experimental results on two real-world datasets demonstrate that the performance of MasterGNN on both air quality and weather prediction tasks consistently outperforms seven state-of-the-art baselines.

\bibliography{ref}
\bibliographystyle{aaai}
	
\end{document}